%% file: paper.tex
\documentclass[letterpaper, 10 pt, conference]{ieeeconf}  

\IEEEoverridecommandlockouts                              

\input{macro.tex}
\usepackage{pifont}

\title{\LARGE \bf
NavForesee: A Unified Vision-Language World Model for Hierarchical Planning and Dual-Horizon Navigation Prediction
}

\author{Fei Liu$^{*}$\quad Shichao Xie$^{*}$\quad Minghua Luo\quad Zedong Chu$^{\dagger}$ \\Junjun Hu\quad Xiaolong Wu$^{\dagger}$\quad  Mu Xu\quad\\
\normalsize{Amap, Alibaba Group \quad} \\
{\tt\small \{lf501161, tenan.xsc, luominghua.lmh, chuzedong.czd, hujunjun.hjj,} \\
{\tt\small  huanlu.wxl, xumu.xm\}@alibaba-inc.com} \\
\thanks{*Joint first authors \quad $\dagger$Corresponding authors}
}

\input{package}

\begin{document}

\let\oldtwocolumn\twocolumn
\renewcommand\twocolumn[1][]{
    \oldtwocolumn[{#1}{
        \centering
        \vspace{-20pt}
        \includegraphics[width=1\textwidth]{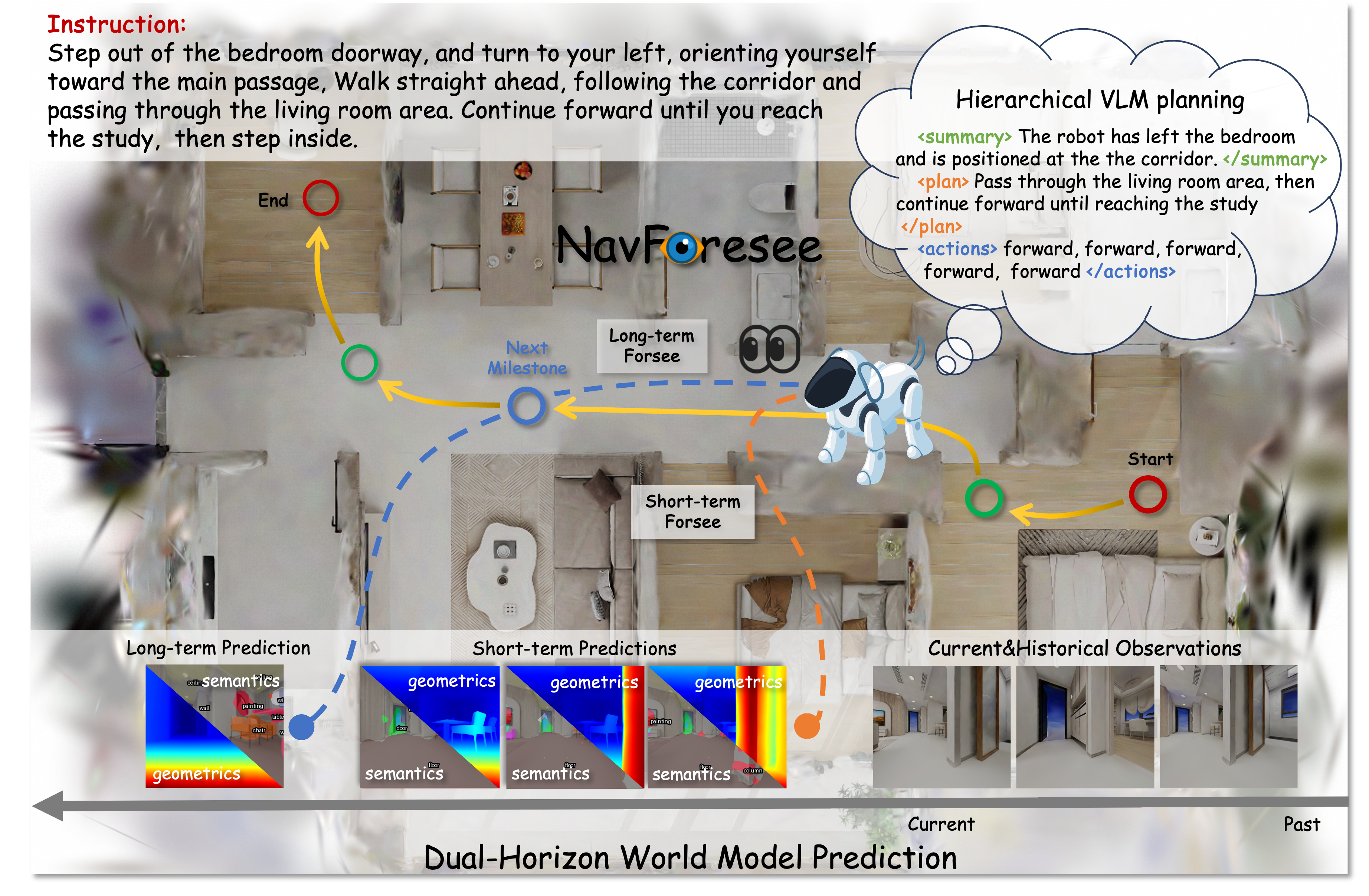} 
        \captionof{figure}{NavForesee integrates hierarchical language planning with dual-horizon predictive foresight. 
                            The planner decomposes instructions into milestone-based sub-goals, while the world model predicts 
                            high-level environmental features for long- and short-term guidance, producing coherent navigation actions.
        }      
        \vspace{15pt}
        \label{fig:navforesee}
    }]
}

\maketitle



\input{section/abstract}

\input{section/intro}

\input{section/related_works}

\input{section/method}

\input{section/experiments}

\input{section/conclusion}

\bibliographystyle{IEEEtran}
\bibliography{IEEEabrv,references}


\clearpage
\setcounter{page}{1}

\setcounter{section}{0}
\oldtwocolumn[
\begin{center}
    \vspace*{1em} 
    {\LARGE \bf Supplementary Material \par}
    \vspace*{1em} 
\end{center}
]

\input{section/supplementary}

\end{document}

%% file: macro.tex
\usepackage[ruled,vlined]{algorithm2e}
\DontPrintSemicolon



%% file: package.tex
\usepackage[table]{xcolor}
\usepackage{titletoc}
\usepackage{tocloft}
\usepackage{lipsum}
\usepackage{url}
\usepackage{graphicx}
\usepackage{subcaption}
\usepackage{listings}
\usepackage[nolist]{acronym}
\usepackage{amsmath}
\usepackage[font={small}]{caption}
\usepackage[export]{adjustbox}
\usepackage{amssymb}
\usepackage{bm}
\usepackage{booktabs}
\usepackage{balance}
\usepackage{color}
\usepackage{dsfont}
\usepackage{esvect}
\usepackage{float}
\usepackage{graphicx}
\usepackage{import}
\usepackage{mathtools}
\usepackage{multirow}
\usepackage{multicol}
\usepackage{flushend}
\usepackage{outline}
\usepackage{paralist}
\usepackage{siunitx}
\usepackage{stackengine}
\usepackage{stmaryrd}
\usepackage{systeme}
\usepackage{soul}
\usepackage{url}
\usepackage{tikz}
\usetikzlibrary{tikzmark}
\usetikzlibrary{calc}
\usepackage[normalem]{ulem}
\let\labelindent\relax
\usepackage{enumitem}
\usepackage{algorithm2e}
\usepackage{scalerel}
\usepackage{makecell}
\usepackage{stfloats}

\usepackage[utf8]{inputenc}
\usepackage{float}

\usepackage{times}
\usepackage{epsfig}
\usepackage{graphicx}
\graphicspath{{./images/}}  
\usepackage{multirow}
\usepackage{amsmath}
\usepackage{amssymb}
\usepackage{bbding}
\usepackage{pifont}
\usepackage{wasysym}

\usepackage{url}
\usepackage{booktabs}
\usepackage{bbding}
\usepackage{utfsym}
\usepackage{fontawesome}
\usepackage{placeins}

\usepackage{xcolor}

\definecolor{citecolor}{HTML}{0071bc}
\definecolor{brightorange}{RGB}{255,225,0} 

\usepackage[pagebackref=true,breaklinks=true,urlcolor=blue,colorlinks,citecolor=citecolor,bookmarks=false]{hyperref}

%% file: section/abstract.tex
\begin{abstract}

Embodied navigation for long-horizon tasks, guided by complex natural language instructions, remains a formidable challenge in artificial intelligence. Existing agents often struggle with robust long-term  planning about unseen environments, leading to high failure rates. To address these limitations, we introduce NavForesee, a novel Vision-Language Model (VLM) that unifies high-level language planning and predictive world model imagination within a single, unified framework.
Our approach empowers a single VLM to concurrently perform  planning and predictive foresight. Conditioned on the full instruction and historical observations, the model is trained to understand the navigation instructions by  decomposing the task, tracking its progress, and formulating the subsequent sub-goal. Simultaneously, it functions as a generative world model, providing crucial foresight by predicting short-term environmental dynamics and long-term navigation milestones. The VLM's structured plan guides its targeted prediction, while the imagined future provides rich context to inform the navigation actions, creating a powerful internal feedback loop of perception-planning/prediction-action. We demonstrate through extensive experiments on the R2R-CE and RxR-CE benchmark that NavForesee achieves highly competitive performance in complex scenarios. Our work highlights the immense potential of fusing explicit language planning with implicit spatiotemporal prediction, paving the way for more intelligent and capable embodied agents.

\end{abstract}

%% file: section/intro.tex
\section{Introduction}
Embodied navigation, a cornerstone challenge in artificial intelligence, has recently witnessed remarkable progress driven by the advent of Vision-Language Models (VLMs) \cite{vicuna2023, LL3DA, GaussianVLM, chat-scene, visualLT}. These models endow agents with the ability to perceive, interpret instructions, and operate in complex environments. Despite these advances, a significant performance gap persists in long-horizon tasks, where agents frequently fail to maintain course, comprehend observations, or make consistently correct decisions. This gap stems from two primary limitations: (1) a planning and memory deficit, as deployable VLMs often have limited context windows and planning capabilities, causing them to get "lost" in the navigation environment\cite{NavR1, DualMemory, GridMM}; and (2) a lack of predictive foresight, as current models are fundamentally reactive and cannot anticipate future environmental states to guide their actions proactively \cite{HNR, ImagineNav, NavMorph}.

Existing research has pursued these challenges on separate fronts. One trajectory enhances VLM reasoning through curated datasets and Chain-of-Thought (CoT) prompting \cite{NavCoT, Aux-think}. The other develops world models to predict future states, informing action planning \cite{CoT-VLA,Ladi-WM}. However, a critical oversight is the disconnection between these paradigms. A VLM-centric agent can suffer from semantic hallucinations, where its plan disconnects from visual reality, while a world model without language guidance can experience semantic drift, its predictions becoming untethered from the instructional goal.

We posit that VLM planning and predictive foresight should not be separate but unified and mutually reinforcing within a single VLM \cite{WordlVLA}. To this end, we introduce NavForesee as in Figure~\ref{fig:navforesee}, a unified model that integrates multi-modal understanding with world model generation. Our approach is inspired by human navigation, which is not a continuous, low-level process but a hierarchical one centered on milestones. Humans typically navigate by heading towards a sequence of meaningful landmarks, largely ignoring the minutiae of the path between them. We argue that an artificial agent should do the same. NavForesee adopts this strategy by operating through two synergistic functions: (1) Hierarchical  Language planning. As a planner, NavForesee generates a high-level plan by summarizing the navigation task into completed sub-instructions, identifying the current sub-instruction, and formulating the next steps as semantic action "trunks." This grounds the agent's planning in the overall instruction. (2) Dual-Horizon Predictive Foresight. As a world model, NavForesee "imagines" the future on two timescales. For long-term guidance, it predicts the key visual features of the environment at the completion of the current sub-instruction—effectively envisioning the next milestone. For short-term execution, it forecasts immediate future features to enhance local awareness, enabling robust obstacle avoidance and better understanding of environmental dynamics.
Inspired by latent-space world models \cite{DINO-Foresight, BackDino, DreamVLA, Ladi-WM}, this prediction deliberately avoids computationally expensive pixel-level generation. Instead, NavForesee forecasts a compact set of high-level features—depth, DINOv2, and SAM features—that capture essential geometric and semantic information as in DreamVLA. The predicted features are fed to an action policy module which is simply an MLP to generate continuous waypoints and flags for arriving or not. By tightly coupling hierarchical  planning with dual-horizon predictive foresight, NavForesee generates coherent, goal-oriented actions, guided by both a long-term vision of its milestones and an immediate awareness of its surroundings. 

We conducted extensive experiments on the R2R-CE \cite{R2R} and RxR-CE \cite{RxR} benchmarks. Training exclusively on the publicly available R2R-CE and RxR-CE datasets, NavForesee demonstrates highly competitive performance, achieving a Success Rate (SR) of 66.2\% and an Oracle Success Rate (OSR) of 78.4\% on the R2R-CE benchmark—comparable to state-of-the-art methods.
In summary, our key contributions are threefold:
\begin{itemize}

\item We propose NavForesee, a VLN framework that unifies vision–language model (VLM) planning with world model prediction for navigation tasks.

\item We introduce a hierarchical language planning paradigm that addresses long-instruction, goal-oriented missions by explicitly tracking mission progress and generating concise textual sub-plans.

\item We design a dual-horizon world model prediction mechanism for both short-term execution and long-term milestone navigation, implicitly forming a perception–planning and prediction–action loop that guides agent behavior.

\end{itemize}

%% file: section/related_works.tex
\section{Related Works}

\subsection{Visual Language Navigation}
Vision-and-Language Navigation (VLN) requires an embodied agent to interpret natural language instructions, perceive visual surroundings, and generate a sequence of actions to reach a specified goal. The advent of large-scale pre-trained  VLMs has catalyzed significant progress, largely superseding earlier methods based on topological graphs \cite{LVER, EGP, TGAL}, top-down semantic maps \cite{SASRA, CML, WSMG}, or instruction augmentation \cite{FlexVLN}. Recent works leveraging VLMs can be broadly categorized into two main paradigms.
The first uses the VLM as a high-level planner, auto-regressively generating action plans \cite{DBM, A2Nav, InstructNav} or textual trajectories \cite{DreamNav}. While strong in reasoning, this step-by-step generation is prone to error accumulation and slow inference. The second employs the VLM as an end-to-end policy, directly mapping inputs to actions. However, this often leads to overfitting on training scenes and underutilizes the VLM's high-level reasoning capabilities.

To bridge the gap between these two approaches, dual-system architectures have been proposed \cite{NavR1, OmniNav}. These models often adopt a "Fast-and-Slow" reasoning paradigm, combining a deliberative "slow" system for high-level reasoning with a lightweight "fast" reactive controller for low-level execution. Reinforcement learning is frequently employed to align the outputs of both systems and bootstrap the learning of coherent reasoning-action patterns. Despite this progress, a fundamental challenge remains: long, complex reasoning chains (e.g., long CoTs) do not always align with the spatial and dynamic realities of the environment. Furthermore, frequent or periodic elaborate reasoning processes may be unnecessary, as human navigation often relies on simpler, high-level semantic plans rather than continuous, detailed deliberation.

\subsection{Navigation World Model}

A world model is designed to learn a predictive model of an environment, forecasting future states from historical observations and optional conditioning information, such as actions or instructions. Predictions can be generated in either raw pixel space or a more compact latent space \cite{BackDino}. The concept has gained significant traction recently, propelled by large-scale video generation models like Sora(\cite{Sora}, which can produce long-term, consistent, even interactive video sequences from text prompts. A key application of world models in robotics is to serve as a simulation engine, allowing an agent to "imagine" the outcomes of different action sequences and evaluate control policies before execution \cite{WordlVLA, DreamVLA}.

In the context of visual navigation, recent works have begun to leverage world models to provide agents with environmental foresight. For instance, NavMorph utilizes a Recurrent State-Space Model (RSSM) to model environmental dynamics in a compact latent space, refining the agent's policy with imagined future states \cite{NavMorph}. Similarly, HNR \cite{HNR} advocates for predicting multi-level semantic features instead of raw pixels, enabling faster and higher-quality imagination to evaluate multiple next-step actions in parallel. Other approaches, like NWM \cite{NWM}, use a controlled video generation model to plan entire trajectories through simulation.
Despite their promise, existing world models for navigation face two primary limitations. First, action-conditioned models that rely on extensive trajectory sampling and evaluation are often computationally prohibitive, rendering them infeasible for deployment on resource-constrained agents. Second, and more critically for our work, prior research has focused almost exclusively on learning environmental dynamics, largely neglecting to integrate this predictive capability with the high-level language reasoning abilities of modern VLMs. This separation leaves a critical gap, which our work aims to address by unifying these two powerful paradigms.
\begin{figure*}[tp]
\centering
\includegraphics[width=\linewidth]{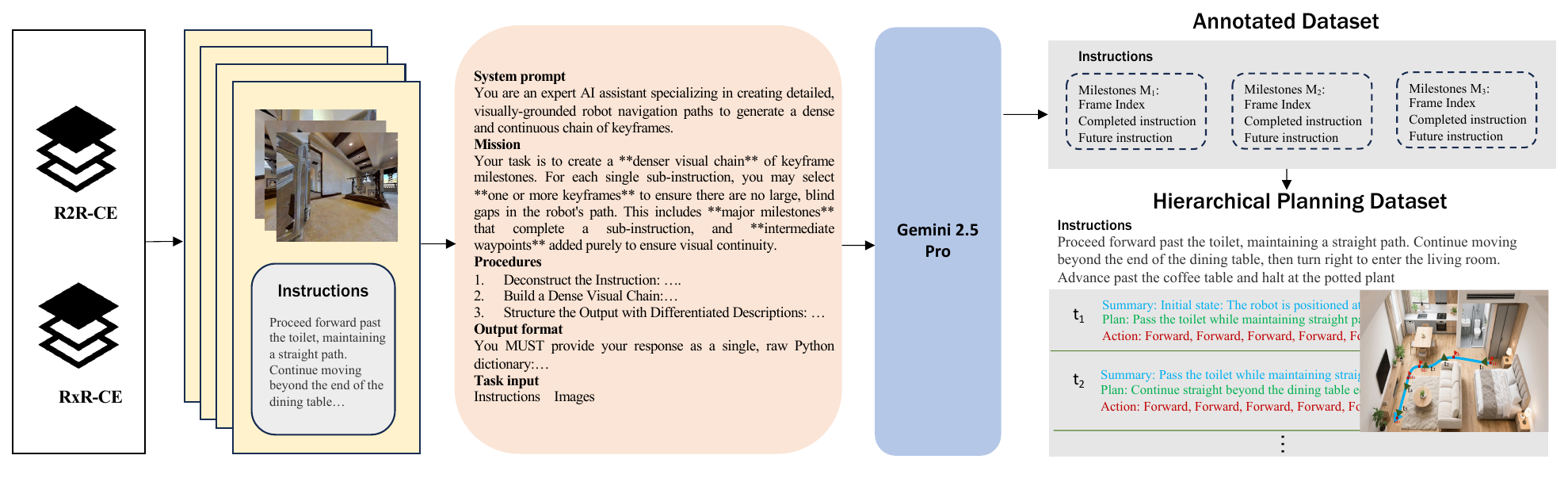}
\caption{VLM-driven hierarchical navigation plan dataset generation. Episodes from R2R-CE and RxR-CE are processed by \texttt{Gemini~2.5~Pro}, which decomposes long instructions into sub-instructions and identifies keyframe milestones. To generation of waypoint-level reasoning labels, waypoints are sampled between milestones annotated with a navigation summary, future plan, and action (\texttt{forward}, \texttt{left}, \texttt{right}, or \texttt{stop}).}
\label{fig2}
\end{figure*}

%% file: section/method.tex
\section{Methods}

\begin{figure*}[tp]
\centering
\includegraphics[width=\textwidth]{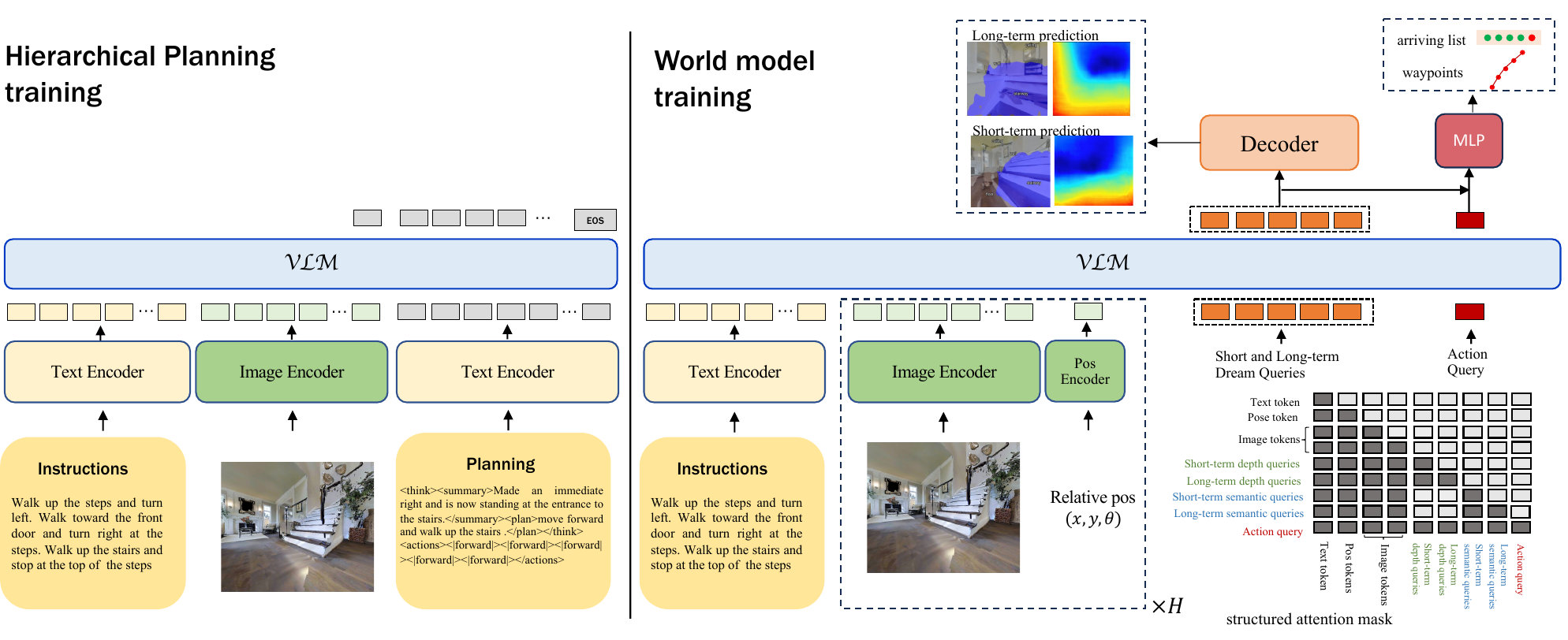}
\caption{ Overall architecture of NavForesee. The model is built on the Qwen2.5-VL-3B-Instruct backbone, integrating two complementary functionalities: (1) VLM-based hierarchical planning and (2) world model-based dual-horizon visual prediction. For hierarchical planning, textual instruction and visual observations are encoded via Qwen’s original multimodal encoders to produce auto-regressive sub-goal plans. For prediction, a position encoder encodes the agent’s relative pose, and short- and long-horizon dream queries (depth and semantic subqueries) are appended to multimodal embeddings. These queries, processed through structured attention, feed lightweight convolutional decoders for environmental predictions and an MLP head for navigation actions.}
\label{fig3}
\end{figure*}

\subsection{Problem Formulation}
We target instruction-guided navigation missions in which an embodied agent must interpret a natural language instruction $l$ and navigate from a given start position to an intended goal location, strictly following the described route. The challenge lies in robustly understanding the instruction, maintaining situational awareness over long horizons, and deciding actions that lead to successful navigation in unseen environments.

At time step $t$, the agent perceives the environment and obtains a panoramic RGB observation $o_t$. It maintains a memory of the past $H$ observations, $O_{t-H:t-1} = [o_{t-H}, \dots, o_{t-1}]$, to support temporal reasoning. The navigation policy produces a sequence of $K$ future waypoints $w_{t:t+K} \in \mathbb{R}^{K\times 5}$, where each waypoint is defined as
\[
w_t = \left[ x_t, y_t, \sin\theta_t, \cos\theta_t, c_t \right],
\]
with $(x_t, y_t)$ denoting planar positions, $\theta_t$ the heading angle, and the binary flag $c_t$ indicating whether a \texttt{stop} action should be triggered. Unless all predicted actions are marked as \texttt{stop}, the agent continuously moves following the generated waypoints.

To solve this problem, we adopt Qwen2.5-VL as our backbone and extend it with two complementary modules. First, we enable \textbf{hierarchical planning} by decomposing the full instruction into sequential sub-instructions, identifying completed ones and predicting the next step under the current context—leveraging the model’s language understanding capabilities and pretraining on our constructed dataset. Second, we integrate \textbf{world model foresight} for predicting short- and long-term environmental changes, enhancing vision–language coherence and yielding more reliable action policies. Together, these capabilities allow the agent to imitate human navigation behaviors, combining explicit language planning with implicit spatiotemporal prediction.

\subsection{VLM-driven Hierarchical Planning Dataset}
We construct a hierarchical language planning dataset specifically for instruction-guided navigation missions, leveraging advanced Vision–Language Models (VLMs) for multi-modal understanding and sequence analysis. Our goal is to provide training data that captures both short-term execution steps and long-term navigation milestones.

As illustrated in Figure~\ref{fig2}, we start from public Vision-and-Language Navigation (VLN) benchmarks—R2R-CE (10k episodes) and RxR-CE (20k episodes)—which provide paired natural language instructions and full image observation sequences. Each episode is processed with \texttt{Gemini~2.5~Pro}, guided by a custom prompt template that specifies the model’s role, defines the mission, outlines analytical steps, and enforces an explicit output format. The VLM systematically decomposes each long instruction into a series of sequential sub-instructions, while identifying a dense visual chain of keyframes representing navigation milestones. For paths involving extended travel or sharp turns, we require the inclusion of intermediate milestones to maintain visual continuity in the generated plan. This hierarchical structure enables downstream world models to better learn both short-term and long-term prediction.

For every annotated episode, the output is standardized to include: the milestone frame index, the textual description of the completed sub-instruction, and the upcoming planned instruction. Post-processing involves filtering incomplete annotations, correcting logical inconsistencies in the VLM outputs, and converting each episode into multiple navigation segments. We sample waypoints along each trajectory, with each waypoint forming the endpoint of a segment between milestones. Each sampled waypoint is assigned a \emph{planning label} comprising: (1) a navigation summary (completed sub-instruction), (2) a future plan (next instruction), and (3) a language action (\texttt{forward}, \texttt{left}, \texttt{right}, \texttt{stop}).

This pipeline produces approximately $1.3$M training samples from RxR-CE and $0.2$M from R2R-CE. To ensure balanced training data, we down-sample over-represented straight-motion cases and augment examples involving stopping actions. The final dataset provides richly annotated, balanced samples for training the hierarchical language planning and predictive modules in NavForesee.

\subsection{Model Architecture}
\noindent \textbf{Overall Architecture}
The overall architecture of NavForesee is illustrated in Figure~\ref{fig3}. We adopt Qwen2.5-VL-3B-Instruct \cite{Qwen2.5} as the backbone. NavForesee is designed to integrate two complementary functionalities: VLM-based hierarchical planning and World model-based visual prediction.
Correspondingly, we define two primary training objectives: hierarchical planning training and world model training. We optimize the hierarchical planning and world model objectives in a unified manner by interleaving their respective training data, ensuring that the model preserves its multi-modal planning ability while simultaneously extending its capability to generate visual features. 
For the hierarchical planning training, textual planning data are directly fed into Qwen for auto-regressive training, leveraging its original encoder components without modification.
For the world model training, we introduce a position encoder and integrate dream queries (covering short/long-horizon depth and semantics) and action queries into the multi-modal input. These are processed via structured attention, where lightweight  decoders transform dream embeddings into environmental features and an MLP predicts navigation actions. More details are  in the supplementary material.

\noindent \textbf{Structured Attention Mask} To maintain a clear separation between short- and long-horizon predictions and to avoid cross-type contamination, each dream query type (depth and semantics) is explicitly decomposed into short-horizon and long-horizon components. As illustrated in Figure~\ref{fig3}, we design a structured attention mask tailored for dual-horizon prediction. Long-horizon predictions naturally depend on short-horizon predictions, using them as guidance to ensure temporal coherence. Mutual attention between depth and semantics queries is masked to prevent cross-modal leakage or unintended feature mixing. In contrast, the action query attends to all available information—including past context and both horizons of dream queries—enabling it to make globally consistent navigation predictions.

\subsection{Dual-horizon World Model Prediction}
Specifically, to enable accurate dual-horizon environmental feature prediction, we employ the world model architecture that serves as guidance for learning the inverse dynamics of a navigation agent. Here, short-term prediction refers to generating forecasts for $k$ steps ahead, while long-term prediction targets navigation milestones, corresponding to an adaptive horizon determined by progress toward the next milestone.

For visual feature prediction within Qwen2.5-VL, we introduce two sets of learnable dream queries, namely the short-term $Q_S \in \mathbb{R}^{L \times d}$ and and long-term $Q_L \in \mathbb{R}^{L \times d}$ to, which extract temporally aligned feature embeddings specialized for prediction at distinct horizons. To enhance the model’s capability in capturing spatial-temporal correlations and learning environmental dynamics, we further integrate position-orientation state embeddings $s_{t-H:t}$ for each input frame through an encoder $h(.)$.
These dream queries are concatenated with textual instruction embeddings $l$ and visual observation sequences $O_{t-H:t}$ and processed by the  Qwen2.5-VL backbone $f(.)$. Specially, 
\begin{equation} \notag
\begin{aligned}
&E_S = f(l, O_{t-H:t}, h(s_{t-H:t}) | Q_S), \\
&E_L = f(l, O_{t-H:t}, h(s_{t-H:t}), Q_S | Q_L) \\
\end{aligned}
\end{equation}
where causal attention masking ensures auto-regressive generation: short-term embeddings are produced first, and long-term embeddings are conditioned on short-term predictions.

We design lightweight decoders to interpret $E_L$ and $E_S$ into predicted depth $d_p $, and high-level semantics $c_p$(e.g. derived from DINOV2, SAM). Short-term predictions correspond to a fixed horizon $k$ whereas long-term predictions adaptively extrapolate over $M_t$ steps, dependent on the agent’s progress toward the next milestone:
\begin{equation} \notag
\begin{aligned}
&p_{t+k} = D(E_S) = [d_p(t), c_p(t)], \\
&p_{t+M_t} = D(E_L) = [d_p(t+M_t), c_p(t+M_t)] \\
\end{aligned}
\end{equation}

The long-term prediction horizon $M_t$ is not rigid or preset. While sub-instructions are learned implicitly during hierarchical planning, the planning-aware hidden states effectively encode the current sub-goal intent. This alignment is achieved through the interleaved training of planning and prediction tasks via shared representations.

\subsection{Predictive Action Policy Learning}
Given two temporally order states $o_t$ and $o_{t+1}$, the intermediate action $\hat{a}(t)$ can be inferred via inverse dynamics. We leverage this principle to learn an action policy conditioned on the instruction $l$, historical visual observations $O_{t-H:t}$ and the dual-horizon predictive latent features $E_S$ and $E_L$ generated by the world model.
To enhance the encoding of task-relevant context for action prediction, we introduce a learnable action query $Q_a$. This query is concatenated with the dream queries and multi-modal input embeddings to form a unified action embedding. The Qwen2.5-VL backbone processes these embeddings to produce the contextual representation for action inference, which is subsequently projected into the action space:
\begin{equation} \notag
\begin{aligned}
&E_a = f(l, O_{t-H:t}, h(s_{t-H:t}), Q_S,  Q_L |Q_a) \\
&\hat{a}_{t:t+k} = M_{inv}(E_S, E_L | E_a)
\end{aligned}
\end{equation}
where $E_a$ is the action embedding and $M_{inv}$ denotes the inverse dynamics model.
Notably, in our action policy learning pipeline, the action embedding $E_a$ is extracted through the Qwen2.5-VL backbone, while action predictions are primarily conditioned on the dual-horizon predictive features, ensuring that decisions are informed by both past observations and forecasted environmental dynamics.

\subsection{Close the Planning/Prediction and Action Loop}
For VLM planning training, we finetune Qwen2.5-VL model based on the constructed dataset in an auto-regressive manner independently to build a powerful model capable of conducting hierarchical navigation.  

For world model prediction and action policy learning, the training tasks are classified depth prediction, semantic feature prediction and action prediction. Depth prediction error $L_d$ is measured using the Scale-invariant Logarithmic Loss (SiLogLoss) at the pix-level level. The semantics feature prediction error $L_c$ and action error $L_a$ are computed using mean squared error (MSE).  The overall training loss $L$ comprise $L_d$, $L_c$ and $L_a$
\begin{equation} \notag
L = \alpha L_d + \beta L_c + L_a
\end{equation}
where $\alpha$ and $\beta$ are weighting hyperparameters balancing the tasks.

%% file: section/experiments.tex
\section{Experimental Evaluation}

\begin{table*}[h]\small
\centering
\caption{\small Comparison with other methods on the Val-Unseen split of R2R-CE and RxR-CE}\label{tab1}
\begin{tabular}{lcccccccccccccccc}
\hline
\textbf{Method} &  \multicolumn{4}{c}{\textbf{Observation}}& \multicolumn{4}{c}{\textbf{R2R-CE Val-Unseen}} & \multicolumn{3}{c}{\textbf{RxR-CE Val-Unseen}} \\
        &  S.RGB  & Pano. & Depth   & Odo   & NE $\downarrow$ & OS$\uparrow$ & SR$\uparrow$ & SPL$\uparrow$ & NE$\downarrow$ &  SR$\uparrow$ & SPL$\uparrow$ \\\hline
HPN+DN* \cite{HPNDN} &         & \Checkmark & \Checkmark   & \Checkmark   & 6.31 & 40.0 & 36.0 & 34.0 & - & - & -  \\
CMA* \cite{CMA*} &         & \Checkmark & \Checkmark   & \Checkmark      & 6.20 & 52.0 & 41.0 & 36.0 & 8.76 & 26.5& 22.1 \\
Sim2Sim \cite{Sim2Sim}&         & \Checkmark & \Checkmark   & \Checkmark      & 6.07 & 52.0 & 43.0 & 36.0 & 8.76 & 26.5& 22.1 \\
GridMM* \cite{GridMM}&         & \Checkmark & \Checkmark   & \Checkmark      & 5.11 & 61.0 & 49.0 & 41.0 & - & -& - \\
DreamWalker* \cite{DreamWalker}&         & \Checkmark & \Checkmark   & \Checkmark & 5.53 & 59.0 & 49.0 & 44.0 & - & -& - \\
Reborn* \cite{Reborn} &         & \Checkmark & \Checkmark   & \Checkmark & 5.40 & 57.0 & 50.0 & 46.0 & 5.98 & 48.6& 42.0 \\
ETPNav* \cite{ETPNav}&         & \Checkmark & \Checkmark   & \Checkmark & 4.71 & 65.0 & 57.0 & 49.0 & 5.64 & 54.7& 44.8 \\
HNR* \cite{HNR}&         & \Checkmark & \Checkmark   & \Checkmark & 4.42 & 67.0 & 61.0 & 51.0 & 5.50 & 56.3& 46.7 \\\hline
AG-CMTP \cite{AG-CMTP}&         & \Checkmark & \Checkmark   & \Checkmark & 7.90 & 39.0 & 23.0 & 19.0 & - & -& - \\
R2R-CMTP \cite{AG-CMTP}&         & \Checkmark & \Checkmark   & \Checkmark & 7.90 & 38.0 & 26.0 & 22.0 & - & -& - \\
Instruc-Nav \cite{InstructNav}&         & \Checkmark & \Checkmark   & \Checkmark & 6.89 & - & 31.0 & 24.0 & - & -& - \\
LAW \cite{LAW}&    \Checkmark &  & \Checkmark   & \Checkmark & 6.83 & 44.0 & 35.0 & 31.0 & 10.90 & 8.0& 8.0 \\
CM2 \cite{CM2}&    \Checkmark &  & \Checkmark   & \Checkmark & 7.02 & 41.0 & 34.0 & 27.0 & - & -& - \\
WS-MGMap \cite{WSMGM}&    \Checkmark &  & \Checkmark   & \Checkmark & 6.28 & 47.0 & 38.0 & 34.0 & - & -& - \\
AO-Planner \cite{AO-Planner}&     & \Checkmark & \Checkmark   &  & 5.55 & 59.0 & 47.0 & 33.0 & - & -& - \\
Seq2Seq \cite{Seq2Seq}&   \Checkmark  &  & \Checkmark   &  & 7.77 & 37.0 & 25.0 & 22.0 & 12.10 & 13.9& 11.9 \\
CMA \cite{Seq2Seq}&   \Checkmark  &  & \Checkmark   &  & 7.37 & 40.0 & 32.0 & 30.0 & - & -& - \\
NA Vid \cite{NaVid}&   \Checkmark  &  &        &  & 5.47 & 49.0 & 37.0 & 35.0 & - & -& - \\
Uni-NA Vid \cite{Uni-NaVid}&   \Checkmark  &  &        &  & 5.58 & 53.5 & 47.0 & 42.7 & 6.24 & 48.7& 40.9 \\
NaVILA \cite{NaVILA}&   \Checkmark  &  &        &  & 5.22 & 62.5 & 54.0 & 49.0 & 6.77 & 49.3& 44.0 \\
Stream VLN \cite{StreamVLN}&   \Checkmark  &  &        &  & 4.98 & 64.2 & 56.9 & 51.9 & 6.22 & 52.9& 46.0 \\
CorrectNav \cite{CorrectNav}&   \Checkmark  &  &        &  & 4.24 & 67.5 & 65.1 & \textbf{62.3} & \textbf{4.09} & \textbf{69.3} & \textbf{63.3} \\\hline
NavForesee(Ours) &     & \Checkmark &        &  & \textbf{3.94} & \textbf{78.4} & \textbf{66.2} & 59.7 & 4.20 & 66.3 &  53.2 \\\hline

\end{tabular}
\end{table*}

\begin{table*}[!h]\small
\centering
\caption{Performance comparison between VLM planning and dual-horizon world model prediction }\label{tab2}
\begin{tabular}{cccccccc}
\hline
Index &  VLM planning & Long-term prediction & Short-term prediction& SR $\uparrow$& OSR$\uparrow$ & NE     $\downarrow$& SPL$\uparrow$\\\hline
1     & \Checkmark    &     \Checkmark       &     \Checkmark       & 66.2       & 78.4              & 3.94  & 59.7\\
2     &  \usym{2717}     &     \Checkmark       &   \Checkmark      & 48.8       & 75.5              & 5.61  & 39.4\\
3     &  \usym{2717}     &     \usym{2717}       &   \Checkmark      & 47.8       & 75.3              & 5.77  & 36.1\\
4     & \Checkmark    &     \usym{2717}      &     \Checkmark       & 58.6           & 76.4              &  4.47 & 50.1\\
5     & \usym{2717}    &     \usym{2717}     &    \usym{2717}       & 52.6          & 67.4              & 5.53  & 46.7\\\hline

\end{tabular}
\end{table*}

We evaluate our model using the Habitat simulator on the R2R-CE and RxR-CE datasets.

\textbf{R2R-CE}  is based on Matterport3D environments and provides fine-grained, step-by-step instructions, allowing for detailed guidance at each navigation step. In the simulator, the embodied agent
can execute turns as small as $15^{\circ}$ and perceives the scene through a $90^{\circ}$ horizontal field-of-view.

\textbf{RxR-CE} is a large-scale, multilingual VLN dataset comprising about 126K human-annotated instructions featuring more complex trajectories. The agent in this setting uses a coarser minimum turn
increment of $30^{\circ}$ and a narrower $79^{\circ}$ horizontal field-ofview, which demands more deliberate movement planning for effective scene coverage.

We report standard metrics: Success Rate (SR), Oracle Success Rate (OS), Success weighted by Path Length (SPL), and Navigation Error (NE).

Training details are provided in the supplementary material.

\subsection{Comparison with State-of-the-Art Methods}
Table~\ref{tab1} reports the performance of the proposed method compared with other approaches on the R2R‑CE and RxR‑CE datasets. Note that methods marked with an asterisk (*) denote discrete waypoint-based approaches. Overall, NavForesee demonstrates superior performance compared to state-of-the-art (SOTA) methods.

Specifically, on the val unseen split of the R2R‑CE dataset, NavForesee achieves SOTA performance by improving SR by 1.1\%, OSR by 10.9\%, and reducing NE by 0.3 m. This improvement can be attributed to the integration of the world model prediction module, which enables the agent to better capture environmental dynamics, avoid obstacles, and explore the surroundings more effectively.

In contrast, NavForesee performs slightly worse than SOTA methods on RxR‑CE, indicating limited generalization to more complex environments. It is worth noting that we  train soly on  NavForesee on R2R‑CE and RxR‑CE datasets, whereas other methods exploit diverse and large‑scale datasets to enhance generalization.
Although NavForesee does not consistently outperform all baselines, it achieves the highest OSR across both datasets. This demonstrates the value of incorporating world model prediction into VLN agents and implies its promising potential for future vision‑and‑language navigation tasks.

\subsection{Ablation Study}
As shown in Table~\ref{tab2}, removing any of the three key modules—hierarchical VLM planning, long-term prediction, or short-term prediction—results in clear performance degradation. The full NavForesee model, which combines all modules, achieves the highest SR (66.2\%), OSR (78.4\%), lowest NE (3.94), and best SPL (59.7\%), validating the benefit of their integration. Without VLM planning, the success rate drops sharply to 48.8\% and the SPL decreases by more than 16 points, reflecting the importance of explicit instruction decomposition and progress tracking for efficient navigation. Disabling long-term prediction also leads to a noticeable reduction in SR (58.6\%) and higher NE, highlighting the role of milestone foresight in providing strategic guidance over extended trajectories. When all three modules are removed, navigation quality deteriorates the most, confirming that planning and both prediction horizons together are crucial for accurate, efficient long-horizon navigation. 

We vary the short-term prediction horizon ($k$) from 3 to 5 (waypoint length) to validate the choice of short-term horizon. Results in Table~\ref{tab3} confirm that matching the horizon to the action space ($k=5$) is optimal.
\begin{table}[h!]
\centering
\caption{Ablation study of Short-term horizon.} 
\label{tab3}
\resizebox{\linewidth}{!}{ 
\begin{tabular}{c|ccc|cccc} 
\toprule
short-term horizon & SR $\uparrow$ & OSR $\uparrow$ & NE $\downarrow$ & SPL $\uparrow$ \\ 
\midrule
k=5 & \textbf{66.2} & 78.4 & \textbf{3.94} & \textbf{59.7} \\
k=4 &  64.1          & \textbf{78.6}          & 4.08          & 54.9 \\
k=3 &  56.5          & 77.2          & 4.85          & 48.5 \\
\bottomrule
\end{tabular}
}
\end{table}

\begin{figure*}[t]
\centering
\includegraphics[width=\linewidth]{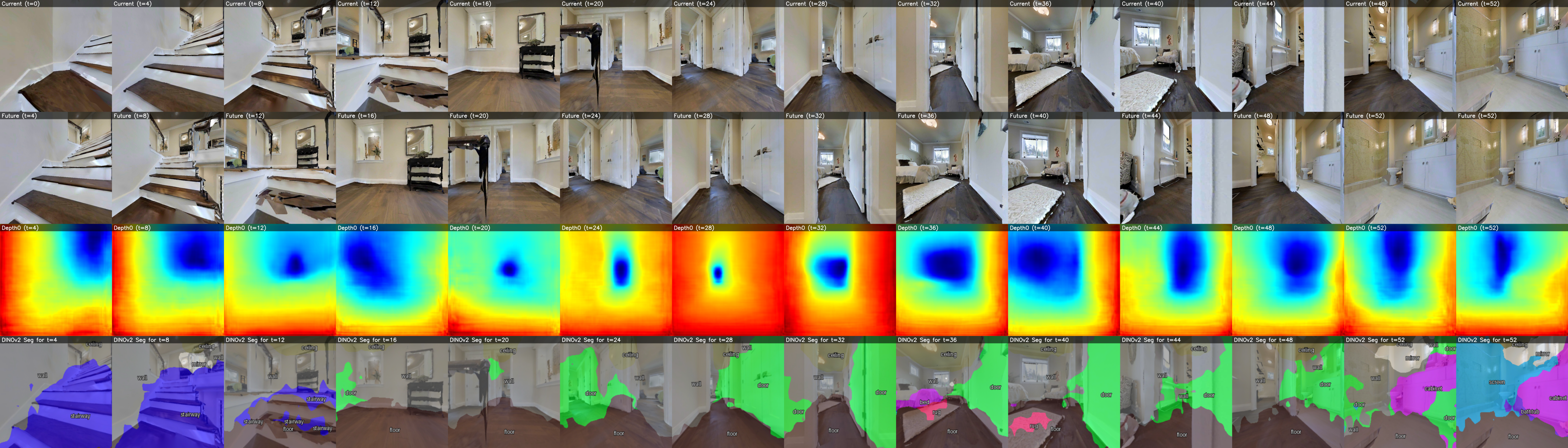}
\caption{Short-term depth and semantics predictions. From top to bottom: frames with timestamps, future ground truth frames with timestamps, future depth prediction for future frames, semantics predictions for future frames. Semantic features are DinoV2 features and visualized by a pretrained segmentation head. Instructions: UP the stairs. Turn to the left and enter into the second open door on the left. Walk towards the foot of the bed. Turn right and enter the open door to the bathroom}
\label{fig5}
\end{figure*}

\begin{figure*}[!t]
\centering
\includegraphics[width=\linewidth]{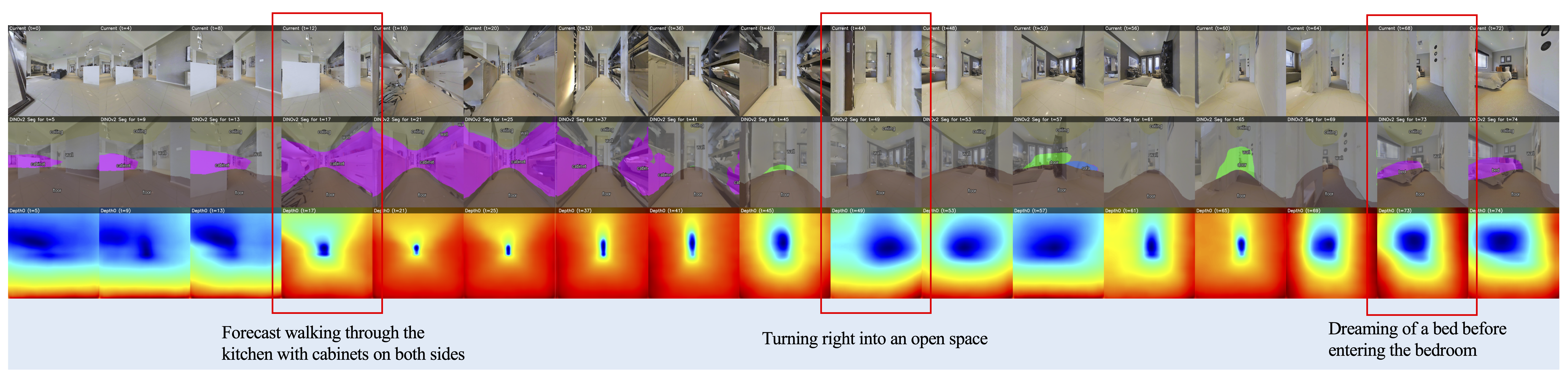}
\caption{NavForesee’s geometric-semantic feature imagination across different motion modes. The model accurately predicts environmental dynamics in straight motion, generalizes effectively to turning scenarios, and infers detailed object geometry and depth distribution from minimal visual input, such as a brief glimpse into a room}
\label{fig6}
\end{figure*}

\subsection{Qualitative Analysis}
Figure~\ref{fig5} illustrates the short-term depth and semantic feature predictions generated by our world model over the course of a complete navigation episode, forecasting up to five future steps. Although the predicted depth maps appear somewhat coarse—owing to the constraints of pixel-level supervised training on R2R-CE and RxR-CE—they nonetheless preserve the scene’s global geometry and spatial layout, faithfully capturing agent movements such as ascending or descending staircases, entering and exiting rooms, and making sharp or gradual turns. This ability to retain high-level spatial coherence despite reduced pixel detail ensures that the model’s predictions remain informative for downstream navigation decisions. The semantics predictions, obtained via a pretrained segmentation head, exhibit strong alignment with ground truth labels, successfully reflecting dynamic environmental changes in synchrony with the agent’s actions. 

Figure~\ref{fig6} further provides detailed examples that showcase NavForesee’s ability to imaginatively anticipate semantic features across diverse motion patterns. In addition to delivering accurate environment dynamics predictions when following a straightforward trajectory, NavForesee demonstrates remarkable generalization by reliably extrapolating future geometric and semantic structures when performing more complex navigational behaviors such as turns. In the final scenario, the agent receives only a brief partial observation—a quick glance into a room—yet the model is able to produce a vivid and coherent internal imagination of the room’s layout. This includes accurately inferring the relative shape and position of the bed, as well as estimating the depth distribution across the room, thus indicating its capacity to reason about unseen spatial regions.

%% file: section/conclusion.tex
\section{Conclusion}
We proposed NavForesee, a vision–language navigation framework that unifies hierarchical language planning with dual-horizon predictive world modeling. By decomposing long instructions into sub-goals and anticipating both short-term dynamics and long-term milestones, NavForesee forms an implicit perception–planning and prediction–action loop.
Experiments on R2R-CE and RxR-CE show strong performance—66.2\% SR and 78.4\% OSR on R2R-CE—comparable to state-of-the-art despite training only on public data. Qualitative results further reveal solid depth and semantics predictions that guide agent decisions in complex scenarios.
These findings highlight the benefit of equipping embodied agents with foresight: by “foreseeing” future states, NavForesee effectively fuses language planning with spatiotemporal imagination to improve visual-language navigation.

%% file: section/supplementary.tex
\section{Implementation Details}
\subsection{Model Architecture}
\noindent \textbf{Base Model} We employ Qwen2.5-VL-3B-Instruct \cite{Qwen2.5} as the backbone of NavForesee. It adopts the Qwen2.5 LLM as its text decoder and integrates a vision encoder. The vision encoder utilizes a Vision Transformer (ViT) architecture to encode visual observations, while the text decoder is responsible for generating the hierarchical planning outputs and action trunk predictions. Detailed descriptions of Qwen2.5-VL can be found in \cite{Qwen2.5}. For hierarchical planning, we directly use the original multimodal encoders and text decoder of Qwen2.5-VL without any modifications. For world model prediction and action policy learning, we introduce a position encoder to represent the agent's relative position and orientation derived from image observations.  
Lightweight decoders transform the dream query embeddings into environmental predictions (depth and semantics), while a simple MLP predicts action outputs (waypoints, orientation estimates, and arrival flags).

\noindent \textbf{Dream Query Design}
Two sets of dream queries (short-term and long-term), along with an action query, are appended to the multimodal embeddings. Each set of dream queries contains depth and semantics subqueries, enabling dual-horizon prediction. We use DINOv2 and SAM features as semantic representations. Thus, there are six query subsets in total—depth, DINOv2, and SAM for both short-term and long-term horizons—with each subset consisting of 64 tokens. The action query consists of a single token dedicated to action prediction.

\noindent \textbf{World Model Decoders}
We design task-specific lightweight world model decoders to transform dream embeddings into depth maps, semantic features, and actions. For depth and semantics predictions, we employ decoder architectures with identical design: dream embeddings and a set of learnable masks are processed by a 2-layer ViT-based decoder to produce predicted features. Additionally, we apply the decoder from VQ-VAE to render depth features into depth maps.

\noindent \textbf{Action Prediction}
The action prediction module takes the action embedding produced by Qwen2.5-VL as input and generates predicted waypoints, orientation estimates, and arrival flags. First, a 2-layer transformer processes the action embedding to capture dependencies on the world model’s dream embeddings. Then, the processed action embedding is passed to the action prediction head, which outputs the final navigation predictions, including waypoints, orientation estimates, and arrival flags. The action prediction head consists of a simple MLP with two linear layers and a ReLU activation in between.

\subsection{Training Details}
We interleave the VLM planning training data and world model training data to jointly train NavForesee. The training batch size is set to 4, and the number of image observations is flexible, up to a maximum length of 20. Depth and semantic features are precomputed and loaded during training. We use the AdamW optimizer with an initial learning rate of $1\times10^{-5}$. Depth and semantics predictions are weighted with $\alpha = 0.25$ and $\beta = 0.3$. The model is trained for a total of 3 epochs on 64 NVIDIA H20 GPUs, with ViT parameters frozen. The fixed short-term prediction horizon is set to 5,  same as the  number of predicted waypoints.

\section{Experimental Evaluations}
\subsection{Hierarchical Planning Evaluation}
\begin{figure*}[t]
\centering
\includegraphics[width=\linewidth]{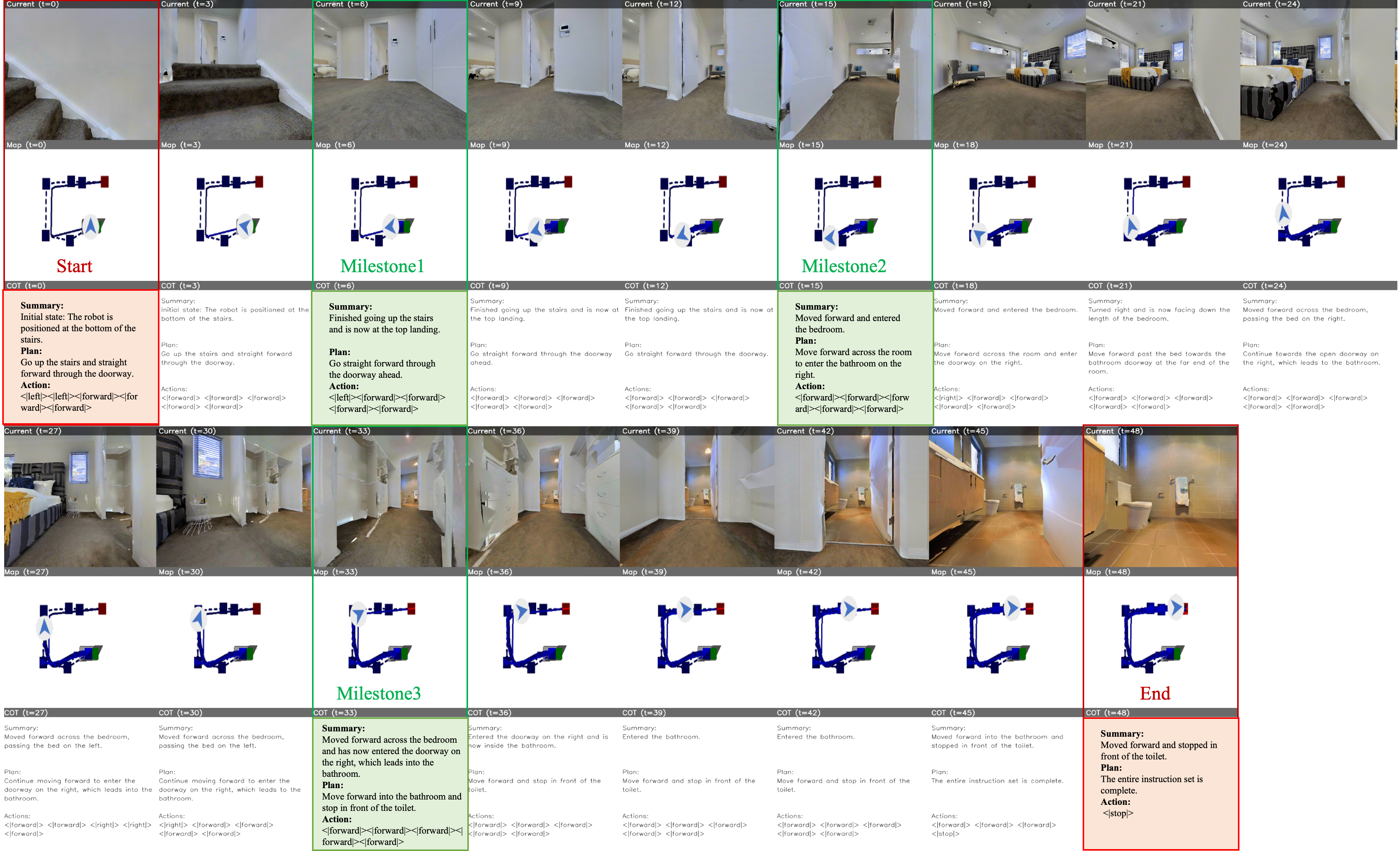}
\caption{Hierarchical planning examples generated by NavForesee for the instruction "Go up the stairs and straight forward the doorway. Turn right, move forward, and enter the doorway on the right. Move forward into the bedroom and stop in front of the toilet". From top to bottom: frames with timestamps, global navigation map, and navigation planning outputs. NavForesee accurately identifies milestones along the route, summarizes completed sub-instructions, and generates the next sub-instruction in accordance with the instruction context.}
\label{fig7}
\end{figure*}

To evaluate the hierarchical planning capabilities of NavForesee, we conduct experiments on the Val-Unseen split of the R2R-CE and RxR-CE datasets. An example is illustrated in Figure~\ref{fig7}. We perform hierarchical planning for each step of an episode. NavForesee generates a navigation summary, plan, and actions strictly following the output format specified in the prompt template. Apart from the initial position, NavForesee consistently identifies milestones along the route, summarizes completed sub-instructions, and formulates the next sub-instruction in alignment with the overall instruction context. This demonstrates that NavForesee effectively leverages its multimodal understanding capabilities to decompose complex navigation tasks into manageable sub-goals, thereby enabling more structured and efficient navigation. Notably, the hierarchical planning module is jointly trained with the world model prediction and action policy learning modules, indicating that NavForesee maintains strong language planning capabilities even when extended with additional functionalities. Furthermore, the hierarchical plans are precise and concise, which greatly benefits subsequent navigation decisions.

\begin{figure*}[t]
\centering
\includegraphics[width=\linewidth]{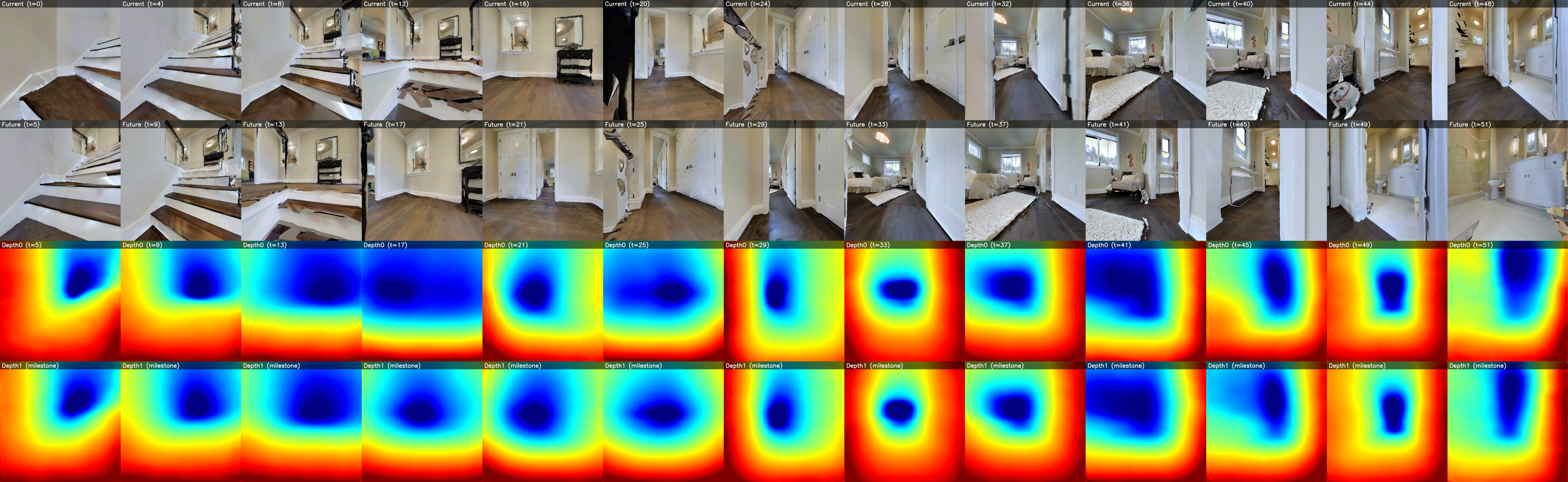}
\caption{Short-term and long-term depth predictions. From top to bottom: frames with timestamps, future ground truth frames with timestamps, short-term depth predictions for future frames, and long-term depth predictions for milestones. Instruction: "Up the stairs. Turn to the left and enter the second open door on the left. Walk towards the foot of the bed. Turn right and enter the open door to the bathroom."}
\label{fig8}
\end{figure*}

\subsection{Short-term and Long-term Prediction Evaluation}
Figure~\ref{fig8} illustrates the short-term and long-term depth predictions produced by our world model over a complete navigation episode. Short-term predictions forecast up to five future steps, whereas long-term predictions extrapolate over an adaptive horizon determined by progress towards the next milestone. Compared to short-term predictions, long-term depth predictions may be less accurate in capturing detailed depth at milestone locations, since milestone positions are unknown during inference. At the beginning of the episode, the long-term predictions effectively capture the scene when the agent ascends the stairs. As the agent approaches the first milestone (the doorway), the long-term predictions degrade slightly, likely due to the increased uncertainty of longer horizons and the absence of explicit milestone information. In such cases, long-term predictions tend to track short-term outputs, because long-term queries can attend to short-term queries. Nevertheless, the long-term predictions maintain the overall scene layout and depth distribution, providing valuable guidance for strategic navigation. This demonstrates that NavForesee's world model effectively anticipates environmental changes over both short and long horizons, enhancing the agent's planning and action capabilities in complex scenarios.

\subsection{Quantitative prediction Quality}
We evaluate prediction quality on R2R Val-Unseen (input at $T$, target at $T{+}5$).  To verify that the model learns temporal dynamics rather than copying inputs, we compare against Identity Baselines that use GT  at $T{+}i$ as predictions for $T{+}5$.
Table \ref{tab4} shows:
1) \textbf{Depth:} Our model outperforms baselines up to $T{+}2$, capturing short-term geometric dynamics.
2) \textbf{Semantics (DINO):} Our model achieves a CosSim of \textbf{0.62}, significantly outperforming baselines up to $T{+}3$.

This indicates strong capability in predicting high-level semantic evolution further into the future.
\begin{table}[h]
\centering
\caption{Prediction Quality. Model (Pred) vs. Identity Baselines (copying GT at $T{+}i$) as predictions for $T{+}5$.}
\label{tab4}
\resizebox{\linewidth}{!}{
\begin{tabular}{c|c|c|ccccc}
\toprule
\multirow{2}{*}{Type} & \multirow{2}{*}{Metric} & \textbf{Model} & \multicolumn{5}{c}{Identity Baselines (GT vs T+5)} \\
 &  & \textbf{Pred} & $T$ & $T{+}1$ & $T{+}2$ & $T{+}3$ & $T{+}4$ \\ 
\midrule
\multirow{3}{*}{Depth} & SSIM $\uparrow$ & \textbf{0.81}    & 0.79   & \underline{0.80}  & 0.82   & 0.84  & 0.88  \\
 & PSNR $\uparrow$ & \textbf{16.39}   & 15.03  & 15.39 & \underline{15.98}  & 17.00 & 19.04 \\
 & LPIPS $\downarrow$ & \textbf{0.28}  & 0.32   & 0.31  & \underline{0.28}   & 0.25  & 0.18  \\ \midrule
Semantic(DINO) & CosSim $\uparrow$ & \textbf{0.62} & 0.47 & 0.50 & 0.53 & \underline{0.58} & 0.67 \\
\bottomrule
\end{tabular}
}
\end{table}

\subsection{Ablation Study on Depth and Semantics Predictions}
We conduct ablation studies to evaluate the individual contributions of depth and semantics predictions in the world model. As shown in Table~\ref{table5}, removing either depth or semantics predictions results in a clear performance drop. The full NavForesee model, which integrates both depth and semantics predictions, achieves the highest SR (66.2\%), OSR (78.4\%), lowest NE (3.94), and best SPL (59.7\%), validating the benefit of their combination. Without depth prediction, the SR drops to 61.8\% and SPL decreases by 4.8 points, highlighting the importance of depth information for spatial reasoning and obstacle avoidance. Disabling semantics predictions leads to an even larger SR reduction (60.0\%) and higher NE, underscoring the critical role of semantic features in recognizing landmarks and guiding navigation. These findings confirm that both depth and semantics predictions are essential for accurate and efficient navigation.

Comparing Table \ref{tab4}  with Table \ref{table5}, we conclude that high-quality semantic foresight is more critical for navigation performance than short-term geometric precision, as it aids in long-term goal anchoring.

\begin{table}[H]
\centering
\caption{Performance comparison between depth prediction and semantics prediction }\label{table5}
\resizebox{\linewidth}{!}{%
\begin{tabular}{ccccccc}
\hline
Index &  Depth  & Semantics  & SR $\uparrow$& OSR$\uparrow$ & NE     $\downarrow$& SPL$\uparrow$\\\hline
1     & \Checkmark    &     \Checkmark          & 66.2       & 78.4              & 3.94  & 59.7\\
2     &  \usym{2717}     &     \Checkmark       & 61.8       & 76.7              & 4.37  & 54.9\\
3     & \Checkmark    &     \usym{2717}         & 60.0       & 76.2              & 4.59  & 52.9\\
\hline
\end{tabular}}%
\end{table}